\documentclass[letterpaper, 10 pt, conference]{ieeeconf}  

\IEEEoverridecommandlockouts                              

\usepackage{graphics} 
\usepackage{graphicx}
\usepackage{soul,color}
\soulregister\cite7
\soulregister\ref7
\soulregister\pageref7
\usepackage{booktabs}
\usepackage{diagbox}
\usepackage{multirow}
\usepackage{amsmath}
\usepackage{balance}
\title{\LARGE \bf
 Are All Point Clouds Suitable for Completion? Weakly Supervised Quality Evaluation Network for Point Cloud Completion
}

\author{Jieqi Shi$^{1}$, Peiliang Li$^{2}$, Xiaozhi Chen$^{3}$ and Shaojie Shen$^{4}$
	\thanks{$^{1}$Jieqi Shi and $^{4}$Shaojie Shen are with Department of Electronic and Computer
Engineering, Hong Kong University of Science and Technology, {\tt\small $\{$jshias, eeshaojie$\}$@connect.ust.hk} 
   $^{2}$Peiliang Li and $^{3}$Xiaozhi Chen are with Dji Co, {\tt\small $\{$peiliang.uav,cxz.thu$\}$@gmail.com}}
}

\begin{document}
	
	\maketitle
	\thispagestyle{empty}
	\pagestyle{empty}

	\begin{abstract}
	
	In the practical application of point cloud completion tasks, real data quality is usually much worse than the CAD datasets used for training. A small amount of noisy data will usually significantly impact the overall system's accuracy. In this paper, we propose a quality evaluation network to score the point clouds and help judge the quality of the point cloud before applying the completion model. We believe our scoring method can help researchers select more appropriate point clouds for subsequent completion and reconstruction and avoid manual parameter adjustment. Moreover, our evaluation model is fast and straightforward and can be directly inserted into any model's training or use process to facilitate the automatic selection and post-processing of point clouds. We propose a complete dataset construction and model evaluation method based on ShapeNet. We verify our network using detection and flow estimation tasks on KITTI, a real-world dataset for autonomous driving. The experimental results show that our model can effectively distinguish the quality of point clouds and help in practical tasks.
	
	\end{abstract}

	\section{INTRODUCTION}
    Current point cloud completion models are often trained in CAD datasets, such as ShapeNet\cite{Chang2015ShapeNetAI}, and applied directly on real-world datasets to assist the downstream tasks\cite{Shi2016RealTimeSI, Wang2021DSPSLAMOO, Giancola2019LeveragingSC}. However, dense and uniform point clouds from CAD datasets are very different from those collected by sensors, especially regarding pattern information, point cloud distribution, noise, and sparsity. Therefore, applying models trained in CAD datasets directly to real-world tasks always brings much trouble. For example, the author of PCN\cite{Yuan2018PCNPC} reported that their completion model achieves an average precision of \textbf{0.009636}(Chamfer Distance) in the ShapeNet test split but obtains a minimal matching distance of \textbf{0.01850} in KITTI. 
    
    To overcome the problem, some works\cite{Shi2022TemporalPC} have tried to augment the point clouds and create noise information while training the completion models, which requires manual assistance in designing the noise. Alternatively, some other works\cite{Najibi2020DOPSLT,Li2021SIENetSI} train the network in CAD datasets and try to refine them in real-world datasets during the training process of downstream tasks. However, due to the lack of ground-truth information, such refinement can not perfectly deal with poor input information in the completion stage but force the subsequent multi-task header to overfit the worse results. Other researchers have tried to use weak supervision or self-supervision to directly complete training on real data\cite{Gu2020Weaklysupervised3S,Mittal2021SelfSupervisedPC}. However, there is still a non-trivial precision gap in the completion results between the models obtained by such methods and models trained with supervision, which makes it difficult for them to satisfy the practical requirements.
    \begin{figure}[t]
		\centering
		\framebox{\parbox{3.2in}{
				\centering

				\includegraphics[scale=0.42]{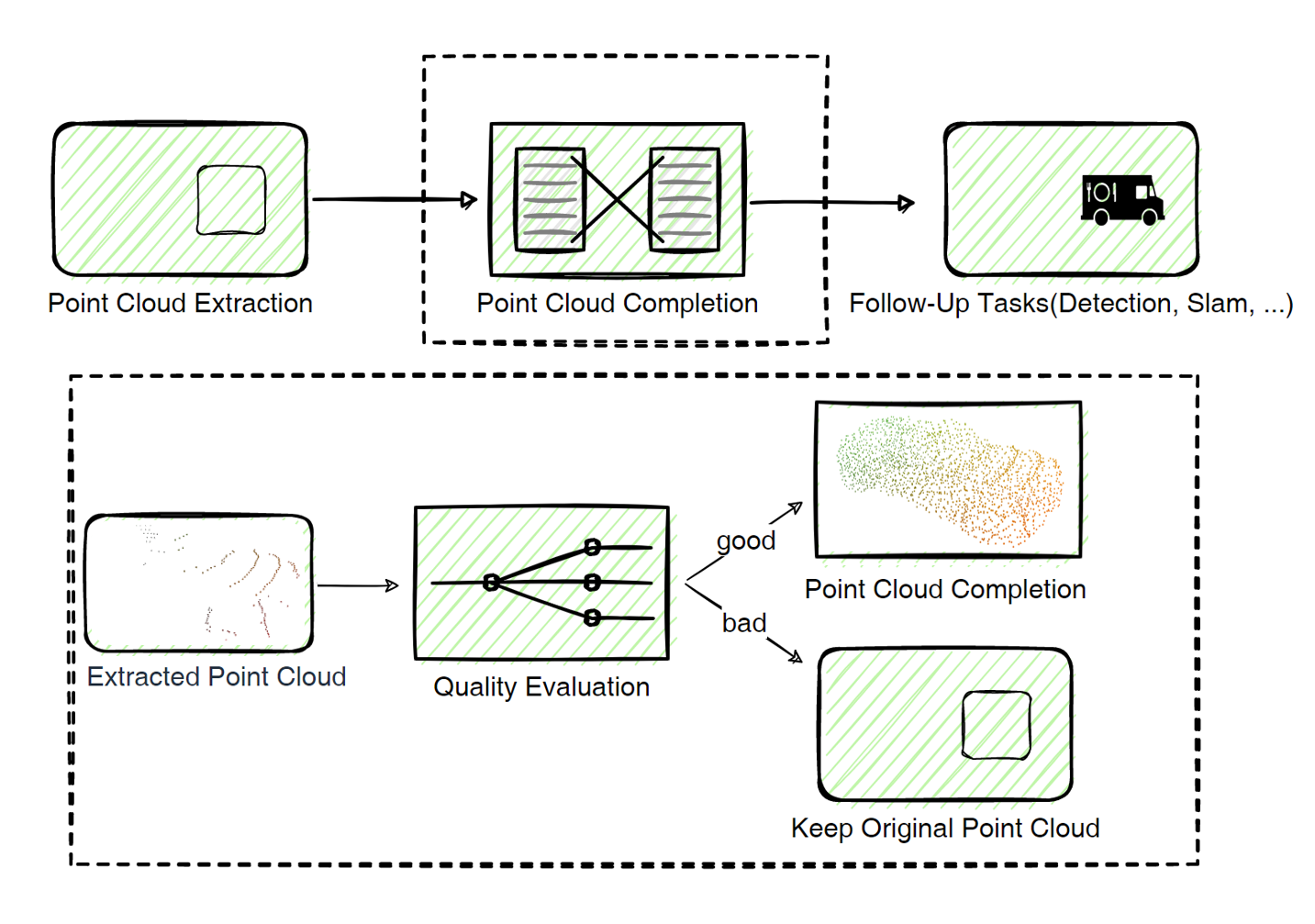}
		}}
		\caption{Illustration of our evaluation pipeline. Upper: Traditional usage of point cloud completion in perception tasks. Lower: Our pipeline.}
		\label{panda}

	\end{figure}	
    
    We summarily refer to the methods mentioned above as model-side enhancement. We can also deal with the problem from the data side. \cite{Yuan2018PCNPC} shows a few failure cases with very high errors in KITTI completion experiments, which significantly bias the mean value. Such a report inspires us to find a filtering method to eliminate the poor data to avoid its influence directly. We emphasized that this method \textbf{does not} fundamentally solve the robustness of the completion task as the model-based method. But we can reduce the impact of noise in the point cloud on subsequent tasks in a simple, lightweight, and minimum-cost mode to meet the accuracy and robustness requirements in practical use.
    
    Three difficulties arise in the design of a point cloud quality network. The first is the establishment of a dataset. Using real-world datasets directly will suffer from the problem of no ground truth data, which makes it hard to tell the model the quality of point clouds in the training stage. However, adding manually designed noise to a CAD dataset will require many statistics and design, and it is challenging to integrate the pattern of the natural scene into the dataset. The second difficulty is the framework of the quality evaluation network. The network must learn from the point cloud to extract the complete shape information and judge its quality. At the same time, it needs to be easy to migrate and fast to implement. Last but not least is the design of the training loss. As mentioned earlier, it is difficult for us to manually design quality judgment rules and tell the network whether the input point clouds are good or not. 
    
    In this paper, we propose a pipeline from the dataset to the network framework and the loss function for the quality evaluation task. We use the SOTA detection models to create a new dataset from ShapeNet. We use an encoder-decoder network inspired by \cite{jieqigcc} to build a simple but effective scoring network. At the same time, we take inspiration from current contrastive learning methods, employing a pre-trained completion network to process the input point cloud and the ground-truth point cloud simultaneously and using the difference as the quality standard of the input point cloud. To demonstrate the effectiveness of our method, we conduct experiments on both the synthetic ShapeNet datset\cite{Chang2015ShapeNetAI} and real-world KITTI dataset\cite{Geiger2012AreWR,Geiger2013VisionMR},  which prove that our model is simple but effective. Our main contributions are:
    \begin{itemize}
        \item [1] We propose a simple but effective network to evaluate the point cloud quality before completion tasks. The network can be inserted into any task as an auxiliary module and helps avoid the influence of noisy data.
        \item [2] We design an indirect dataset construction method to simultaneously utilize the pattern information of real datasets and the ground-truth of CAD datasets.
        \item [3] We propose a quality evaluation standard that uses pre-trained networks to evaluate point clouds to avoid complicated manual rules design.
    \end{itemize}

    \begin{figure*}[ht]
		\centering
		\vspace{2em}
		\framebox{\parbox{6.6in}{
				\includegraphics[scale=0.21]{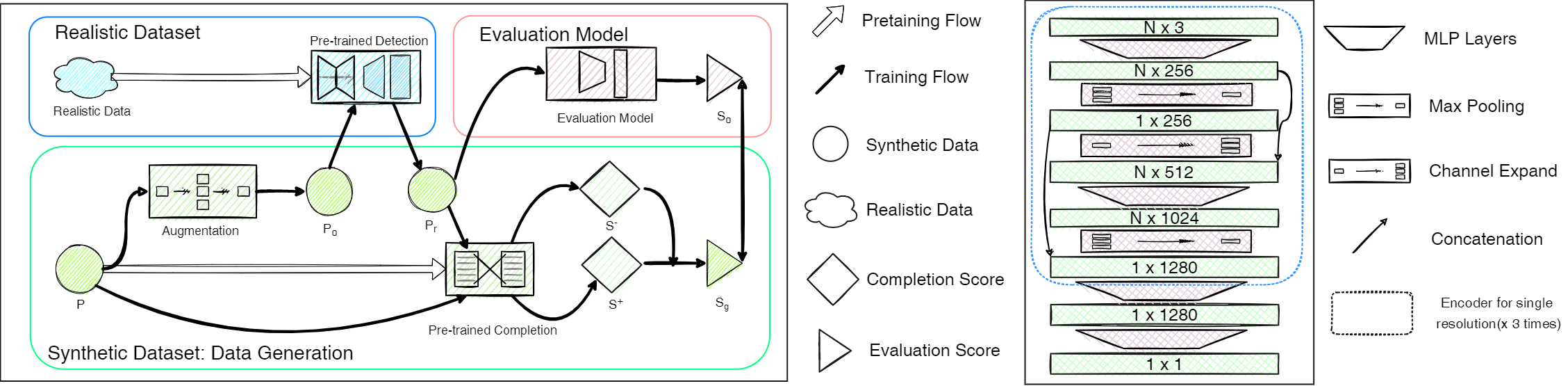}
		}}
		\caption{Details of our framework. Left: The operation pipeline of the system. Different modules are marked with different colors. Right: The detailed structure of our evaluation network. P, $P_o$, $P_r$ corresponds with input PC, augmented PC and PC from dectction models(Section III.A), and $S^+$, $S^-$, $S_g$ are score for positive, negative and GT samples(Section III.C).}
		\label{framework}
		\vspace{-2em}
	\end{figure*}

	\section{Related Work}
	
    \subsection{Deep-learning-based 3D Shape Completion}
    The 3D shape completion task has become a hot research topic. Researchers have deeply studied various forms of input data, including point clouds\cite{Wen2020PMPNetPC,Yin2018P2PNET,Pan2021VariationalRP}, voxels\cite{Dai2017ShapeCU,Xie2020GRNetGR,Stutz2018Learning3S} and implicit fields\cite{Gao2020LearningDT,Thai20203DRO,Liu2020DISTRD}. We can divide the completion task into paired and unpaired approaches based on the training method. Paired methods are the traditional supervised training method. In the network training process, the network accepts both incomplete point clouds and ground-truth pairs to guide the completion process. These methods require detailed data, so they are usually trained on CAD simulation datasets and tested on real data. Researchers have mainly focused on using the network to extract features better to obtain more accurate results. For example, GCC\cite{jieqigcc} and ECG\cite{Pan2020ECGEP} use graph networks to enhance the relationship between points and retain the characteristics of local point clouds. PMP-Net\cite{Wen2020PMPNetPC} uses gate structures to memorize the movement direction of points and to pair the input and output points one by one. VRC-Net\cite{Pan2021VariationalRP} and Snowflake-Net\cite{Xiang2021SnowflakeNetPC} use different attention structures to enhance the features and obtain better shapes. GR-Net\cite{Xie2020GRNetGR} and VE-PCN\cite{Wang2021VoxelbasedNF} turn back to voxels and utilize the uniform voxels for better shapes.
   
   Other researchers have used unsupervised or weakly supervised methods to train completion tasks to better adapt to real-world datasets. For example, AML\cite{Stutz2018Learning3S} and Pcl2Pcl\cite{Chen2020UnpairedPC} directly learn the correspondence or distance between partial and complete point clouds in the latent space, while Cycle4Completion\cite{Wen2021Cycle4CompletionUP} extends this process to a bi-directional matching. MFM-Net\cite{Cao2021MFMNetUS} further applies multi-stage matching to align features and enhance the features with attention modules. Besides these feature-based matching methods, \cite{Gu2020Weaklysupervised3S, Tulsiani2018MultiviewCA} and \cite{Insafutdinov2018UnsupervisedLO} propose to use the consistency between multi-view point clouds and recover the shape and poses at the same time. Based on this idea of consistency, \cite{Mittal2021SelfSupervisedPC} artificially creates occlusions on point clouds and supervises only the parts with ground-truth values to carry out self-supervised training on real data. These methods enable the completion task to be trained and tested on real data to learn the real pattern better. However, unpaired methods usually bring a poor completion effect, and the accuracy is more than one order of magnitude lower than that of supervised learning. Our work aims to keep the accuracy of supervised training methods while achieving better operation in real-world datasets. We do not modify the model itself but select the appropriate data to avoid the impact of noise data on the accuracy of the whole system. 

    \subsection{3D Shape Completion in Practice}
    With the maturity of completion technology, more and more researchers have begun to discuss the application of completion in practical tasks. \cite{Xu2021SPGUD} and \cite{Xu2021BehindTC} divide point clouds into voxels, mask the foreground voxels and complete the occluded parts before or within the detection pipeline. They successfully combine the completion task with the 3D detection task, effectively improving the detection results' accuracy. Based on point clouds, \cite{Najibi2020DOPSLT} trains the completion model in CAD datasets and uses the shape prior to refine the training process of a 3D detection pipeline. \cite{Li2021SIENetSI} combines the voxel-based proposal generation with point-based completion, completing the objects within proposals to enhance the refinement stage. Although most of the completion operations rely on a more accurate foreground segmentation or pre-detection, the above methods prove that the completion and detection tasks can supervise and improve each other and can help in practical fields.
    
    Besides 3D detection, \cite{Giancola2019LeveragingSC} combines a Siamese network with point cloud completion models and proposes the first 3D Siamese tracker based on point clouds instead of images. \cite{Pang2021ModelfreeVT} operates point cloud completion in a traditional manner, obtaining the complete shape by accumulating point clouds of different frames and using the point cloud registration error for the pose correction of the tracking system. \cite{Wang2021DSPSLAMOO} integrates shape completion into the traditional SLAM system, using sparse key-points created by SLAM pipelines to generate the complete shape of objects, and optimizes the shape and object poses together in a bundle-adjustment module. What's more, some researchers are trying to use application to assist the task of view path planning planner\cite{Zhang2021ContinuousAP,Song2022ViewPP,Hepp2019Plan3DVA}.
    
    The system usually does not adopt the machine learning model in tracking or SLAM tasks. Researchers often add completion models to specific modules to enhance the system itself. However, in this case, the data quality may greatly impact the actual process, and the noisy inputs will decline overall quality. Because of the poor quality of single-frame point clouds, \cite{Shi2022TemporalPC} attempts to fuse multi-frame point clouds by using a gated recurrent unit(GRU) structure and attention modules and optimizes the completion result frame by frame. However, this method can not automatically judge how many frames of point clouds are required to meet the system's requirements, so there are still many deficiencies in practical use. In this work, we aim to make judgments from the perspective of data, distinguish the parts of point cloud data that are suitable for and unsuitable for completion, and reduce the noise caused by completion errors to facilitate the application of completion tasks in practical systems.

	\section{Method}
	Our quality assessment network should meet several requirements:
\begin{itemize}

\item[1. ] It should adapt to the patterns of both CAD and real-world datasets to facilitate the smooth migration of models trained on simulated data to real tasks.
\item[2. ] In addition to the ability to encode the whole shape, the model should be easy and fast to implement so that it can be used as a pilot module in practical applications.
\item[3. ] The network output should be a clear indicator, such as a clear score, to intuitively evaluate the quality.
\end{itemize}
Targeting these three requirements, we design the pipeline from three aspects: datasets, model framework, and training loss. We introduce each module separately to facilitate understanding and then describe our overall system process.
    
    \subsection{Dataset Preparation}
    Considering that real data lacks the ground-truth shape and using weakly supervised or unsupervised training methods will lead to significant performance degradation, we construct our dataset based on CAD data. The real data for completion usually comes from the object point cloud generated by three-dimensional detection. Thus, there are many defects, such as the pose disturbance error caused by 3D frame offset, the interference of the ground and other factors, and the parts of point clouds a detection model tends to crop. Although we can enhance the simulated data through artificial noise and disturbance, ensuring that the created point cloud conforms to the real pattern isn't easy.
    
    Luckily, the past research has given us some inspiration. In \cite{Najibi2020DOPSLT} and \cite{Li2021SIENetSI}, a completion network trained on ShapeNet was added to the detection framework to enhance the coarse results. The completion model encodes the shape space of the CAD dataset and uses the prior in real data. The success of these works shows that the shape space of CAD data is similar to that of real data in a large range. Therefore, we believe that we can create a dataset that includes real situations by enhancing the CAD dataset and fusing some real features. Moreover, we consider reversing the process of real applications: training the model in CAD data and using in real data. We can prepare a machine learning model on real data, use it to remember some of the data patterns, and apply this model to CAD datasets for enhancement. Such a method creates a simulation dataset based on CAD data but containing real data characteristics. 
    
    We first build a fundamental 2.5D image set from ShapeNet following PCN\cite{Yuan2018PCNPC} with the ground-truth point clouds. After that, we make four enhancements to the basic point clouds. First, we randomly sample the input points between 128 and 2048 points. Second, we add random pose disturbance to the objects, such as rotating at any angle on the ground plane or translating in any direction. Third, we scale the normalized object in the range of 0.8--1.2. Forth, we randomly select some objects and the ground plane from KITTI to simulate a real scene. These enhancements help us to add noise to clean CAD models and mimic real data collected by lidar sensors.

    We denote the augmented input point cloud with such augmentation as $P_o$. For the ground-truth point clouds, we directly use the original ground truth sampled from synthetic datasets and denote it as $P_g$. At the same time, we use a two-stage detection model trained on real-world datasets to crop out the point clouds within the region-of-interest(ROI) boxes with the top K(K=5 in our experiments) scores. We denote the point cloud generated by the ROI as $P_r$. Since the detection model is trained on real datasets, we believe that it has remembered some real patterns of lidar point clouds, and can introduce such features to our CAD dataset by the detection operation. In short, we can compare this detection model to the backbone model of pre-training, and use the detection method to provide the feature information of real data for subsequent tasks.
    
    \begin{table*}[htbp]
	\centering
	\vspace{2em}
	\caption{Minimal Matching Distance($CD * 10^{-6}$) before and after Filtering}
	\begin{tabular}{c|c|c|c|c|c|c|c|c|c}
		\hline
		 & & Average & Seq 4 & Seq 8 & Seq 10 & Seq 14 & Seq 18 & Seq 20 & Seq 23\\ \hline
		 \multirow{3}[4]*{PCN} &  w/o Filtering & 1.81 & 2.09 & 1.80 & 1.69 & 1.68 & 1.91 & 2.07 & 1.99\\ 
		 & Score 0.5 & 1.77 & 2.01 & 1.75 & 1.70 & 1.67 & 1.98 & \textbf{1.82} & \textbf{1.85}\\
		 & Score 0.9 & \textbf{1.75} & \textbf{1.54} & \textbf{1.71} & \textbf{1.60} & \textbf{1.63} & \textbf{1.95} & 1.96 & 2.08\\ \hline
		  \multirow{3}[4]*{TopNet} & w/o Filtering & 1.86 & 1.92 & 1.76 & 1.81 & 1.76 & 1.83 & 1.89 & 2.04\\ 
		 & Score 0.5 & 1.83 & 1.95 & \textbf{1.71} & 1.81 & 1.78 & 1.81 & 1.89 & 1.98 \\
		 & Score 0.9 & \textbf{1.74} & \textbf{1.65} & 1.74 & \textbf{1.72} & \textbf{1.70} & \textbf{1.79} & \textbf{1.74} & \textbf{1.91}\\ \hline
		  \multirow{3}[4]*{GCC} & w/o Filtering & 1.67 & 1.72 & 1.72 & \textbf{1.63} & 1.55 & 1.64 & 2.10 & 1.83\\ 
		 & Score 0.5 & 1.65 & 1.71 & 1.69 & 1.65 & 1.55 & 1.67 & 2.10 & 1.77\\
		 & Score 0.9 & \textbf{1.60} & \textbf{1.50} & \textbf{1.66} & 1.69 & \textbf{1.39} & \textbf{1.60} & \textbf{1.75} & \textbf{1.75}\\ \hline

	\end{tabular}
	\vspace{-2em}
    \end{table*}	    
    
	\subsection{Network Selection}
	The primary consideration for our network is to have enough ability to evaluate the overall quality of the point cloud shape. Considering that the ultimate goal of our quality assessment is to serve the shape completion models, the intuitive method is to adopt the original encoder network from the SOTA completion methods and change the decoder for the generation to a classification task after obtaining the feature information. Fortunately, most existing completion networks are based on an encoder-decoder structure.
	
    In addition to the feature extraction capability, we also need to consider the efficiency of the network, including running time and memory occupation. Considering that we aim for the quality evaluation module to be easily insertable into any task as an auxiliary task to filter the point clouds, the network needs to use as little time and memory as possible. Therefore, we exclude the voxel-based processing method to avoid the excessive memory expenditure caused by the improvement of the resolution and choose to process the input directly based on the point cloud. 
    
    Based on the above considerations, we adopt a similar encoder structure to those in \cite{jieqigcc} and \cite{huang2020pfnetpf}. We sample the input point clouds on three resolutions. For each resolution, we first use stacked MLP units to expand the dimensions from 3 to 256, get $f_0$, and apply max pooling for a global feature $g_0$. We then expand $g_0$ to the same size as $f_0$ and concatenate them together for a feature $f_1$. We again use MLP layers to expand the dimensions of $f_1$ to 1024 and pool for another global feature $g_1$. We then follow \cite{huang2020pfnetpf} to concatenate $g_0$ and $g_1$ as the output of each resolution, and use another set of MLP layers to get the overall feature of the shape $(N, 1), N=1280$. The shape code is converted into a single score $S_o$ and normalized to 0--1.
    
    \subsection{Quality Evaluation Standard Design}
    The biggest problem in designing the evaluation network is the design of the training loss. As we mentioned earlier, multiple factors make it challenging to complete the point cloud. Too concentrated a point cloud may result in a lack of general information. At the same time, too few point clouds may lead to difficulty in judging the type of object, and noise impact may bring about generation failure, etc. At the same time, a point cloud that seems difficult to complete in the human eye's judgment is likely to meet the machine's requirements fully. In Fig. 3, we show cases where the point clouds can be completed perfectly. Though the point cloud looks pretty sparse and hard to distinguish, it contains all the features required for completion, such as the approximate range, shape, and trend of the vehicle, and can be used to provide complete and smooth completion results. Therefore, we hope to find a way to use deep learning models to automatically judge the quality of a point cloud rather than manual annotation.
    
    \begin{figure}[ht]
		\centering
		\framebox{\parbox{3.2in}{
				\includegraphics[scale=0.49]{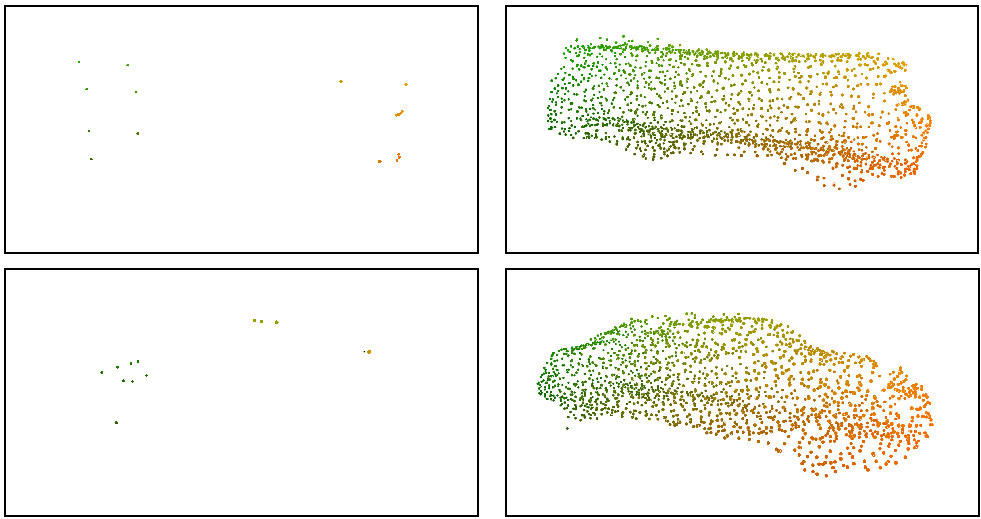}
		}}
		\caption{Point clouds that can hardly be recognized yet satisfy the completion requirements(with sparse points at the front and rear of the vehicle).}
		\label{example}
	\end{figure}
    
    The scoring task can not be described as a yes-or-no binary classification task but is instead to give a confidence within a continuous interval. Thus, it isn't easy to define the task using the popular self-supervised methods. However, we can evaluate the output results with the help of the trained SOTA completion model. Back to our original purpose, the evaluation network is used to filter out the point clouds suitable for the completion task, which means that we can judge the quality of the input point cloud through the quality of the completion results.
    
    Inspired by contrastive learning, we take three samples as a group. A positive sample $P^{+}$, that is, the output result corresponding to an excellent input point cloud, and a negative sample $P^{-}$, the point cloud to be filtered, together with the corresponding ground-truth point clouds $P_g$, are treated abstractly as a set of data. Through the completion network $F$, we can get two results, $F(P^{+})$ and $F(P^{-})$, both of which have a similarity to $P_g$. Obviously, $similarity(F(P^+), P_g) \gg  similarity(F(P^-), P_g)$. Therefore, we can use the proportion of similarity to judge the difference between samples. In our practical use, we choose the Chamfer distance, 
   the most popular function to measure the completion accuracy, as the evaluation index, with the original partial point cloud $P$ as the positive sample and the cropped point clouds $P_r$ as negative samples. The quality standard is defined as
   
   \begin{equation}
   \begin{aligned}
       S^+ &= CD(F(P), P_g),\\
       S^- &= CD(F(P_r), P_g),\\
       S_g &= max(1.0, \frac{S^+}{S^-}).
    \end{aligned}
   \end{equation}
	Since CD decreases with the increase of similarity, the larger $S_g$ is, the better the cropped point cloud is. 
	
    \subsection{Loss Function}
	Although the establishment of datasets and the design of quality standards are complex, the final function of the network is only to obtain a score in the range of 0--1. Therefore, we use the simple Huber loss as the final training loss:
	\begin{equation}
	E = Huber(norm(softmax(S_o) - S_g), delta=2.0)
	\end{equation}
	
	\subsection{Operation Process}
	For a set of data pairs, $P, P_g$, generated on ShapeNet, where $P$ means 2.5D point cloud data, and $P_g$ is the ground-truth point clouds of 2048 points. We first obtain $P_o$ through data enhancement. Next, we feed $P_o$ to the pre-trained detection model to get ROI areas and crop out point clouds $P_r$ within the ROI.
	
	During training, the $(P, P_r, P_g)$ groups are then put into our evaluation system. The quality evaluation network accepts $P_r$ and gives out a predicted score $S_o$. Meanwhile, we use a pre-trained completion model to process $P_r$ and $P$, and calculate the CD loss between $P_g$ and the two completed models for $S^+$ and $S^-$. The division of these two scores is then treated as the ground-truth score $S_g$. We finally calculate the Huber loss between $S_o$ and $S_g$.
	
	We only use the real point cloud $P_r$, and get a quality score $S_o$ during inference. By setting a threshold for $S_o$, we can quickly eliminate input point clouds that do not meet the quality evaluation criteria and avoid noise impact.

    \begin{table}[htbp]
	\centering

	\caption{Inference Time of the Per-Object Quality Evaluation Network(ms) N: Point Number B: Batch}
	\setlength{\tabcolsep}{4mm}{
	\begin{tabular}{c|c|c|c|c|c}
		\hline 
		 \diagbox{N}{B} & 1 & 4 & 8 & 16 & 32\\ \hline
		 128 &  3.53 & 1.49 & 1.06 & 0.79 & \textbf{0.59}\\ 
		 256 & 3.29 & 1.52 & 1.07 & 0.80 & \textbf{0.63}\\ 
		 512 & 4.02 & 1.58 & 1.12 & 0.81 & \textbf{0.71}\\
		 1024 & 5.16 & 1.90 & 1.28 & 0.88 & \textbf{0.70}\\ 
		 2048 & 5.69 & 2.04 & 1.30 & 0.96 & \textbf{0.69}\\ \hline
	\end{tabular}}
	\label{time}
	\vspace{-1em}
    \end{table}

    \begin{table*}[htbp]
	\centering
	\vspace{1em}
	\caption{Detection Results on Kitti Val Dataset(Car AP 0.7, 0.7, 0.7)}
		\setlength{\tabcolsep}{5mm}{
	\begin{tabular}{c|c|c|c|c|c|c|c}
		\hline 
		 Modality & Type & \multicolumn{3}{c|}{PV-RCNN Box(R11)} & \multicolumn{3}{c}{Point-RCNN Box(R40)}  \\ \hline
		  \multirow{3}[3]*{BEV AP}& w/o Filter & 86.50 & 79.93 & 75.97 & 88.91 & 84.55 & 79.64 \\
		  & Random Box & 89.77 & 86.36 & 85.18 & 89.98 & 84.86 & 79.81 \\ 
		  & Dense Box & 89.08 & 84.79 & 84.54 & 88.41 & 85.71 & 84.7 \\
		  & Dense Two Box & 89.44 & 87.04 & 86.79 & 89.37 & 86.92 & 85.06 \\
		 & w. Filter & \textbf{90.11} & \textbf{87.90} & \textbf{87.54} & \textbf{90.18} & \textbf{87.86} & \textbf{85.46}\\ \hline
		 \multirow{3}[3]*{3D BBOX}& w/o Filter & 83.52 & 71.01 & 71.17 & 85.82 & 77.27 & 74.88\\
		 & Random Box & 88.73 & 78.48 & 77.58 & 87.65 & 77.49 & 74.75 \\
		 & Dense Box & 87.69 & 76.77 & 76.48 & 77.58 & 74.70 & 73.21 \\
		 & Dense Two Box & 88.13 & 78.35 & 78.11 & 83.22 & 76.81 & 75.26 \\
		 & w. Filter & \textbf{89.34} & \textbf{83.90} & \textbf{78.78} & \textbf{89.21} & \textbf{78.85} & \textbf{77.94} \\\hline
	\end{tabular}}
	\label{detection_kitti}
	\vspace{-2em}
    \end{table*}	    
	
	\section{Experiments}
	\subsection{Training Details}
	In our operation, we use the PV-RCNN\cite{Shi2020PVRCNNPF} model pre-trained on KITTI, published by the authors, as the detector for creating datasets. We then refine a GCC\cite{jieqigcc} model in the created dataset using the augmented $(P_o, P_g)$ pairs to adapt to the point cloud data with partial noise. The trained GCC model is subsequently used to calculate the ground-truth value of quality evaluation. 
	
    To reduce the data processing time in the training process, we first generate a training and test dataset with a fixed size and combine the $(p, p_o, p_r, p_g)$ group with the pre-calculated ground-truth value $S_g$. During the quality evaluation network training, we use the Adam optimizer and train the network on one GTX 1080 Ti GPU with batch size 32 for 12 epochs. We use the same model for all the following experiments without further refinement.
    
    \subsection{Accuracy}
    The Minimal Matching Distance(MMD) is proposed as an evaluation criterion by PCN\cite{Yuan2018PCNPC} in its experiment section. It is the calculation, using the Chamfer distance, of the shortest distance between the completed point cloud and the point clouds of the same category in the training dataset, and can be used to measure the rationality of the shape of the completion result. We use a completion model trained on ShapeNet to complete cars from the KITTI dataset and calculate the MMD with the ground-truth point clouds in ShapeNet. By checking the MMD of all vehicles against that of the point clouds accepted by the evaluation model, we can quickly determine whether our quality evaluation model can distinguish between a point cloud suitable for completion and a point cloud that may become noise.
   
   We use PCN\cite{Yuan2018PCNPC}, ToPNet\cite{Tchapmi2019TopNetSP} and GCC\cite{jieqigcc} as three benchmarks to test the average MMD of point clouds before and after filtering. We follow previous works, extract the point clouds within bounding boxes, normalize and rotate them to the forward dataset, and feed them to our evaluation pipeline. Here we use a SOTA detection model instead of ground-truth boxes to mimic real-world settings. Considering that our training dataset is created based on the PV-RCNN model, we use Point-RCNN for bounding boxes to avoid over-fitting brought by detection models. Our benchmark models are pre-trained on the clean ShapeNet dataset.
   
   We follow PCN and do our experiments in all 28 sequences from KITTI tracking test split. In Table I, we report the changes in the average MMD and select representative sequences. Due to the significant difference in point cloud quality between sequences, we also show the results of some individual sequences in addition to the average results. On average, our indicators show slight improvement. But on some representative sequences, filtering brings a large degree of performance improvement. For example, in Seq. 4, when the threshold is set to 0.9, our evaluation network filters two-thirds of the vehicle point clouds(421 to 125) and brings a 25$\%$ MMD increase at the same time. However, in the sequence with good data quality, such as Seq. 14, the improvement is very slight, even though our threshold also filters out a large portion of the point clouds(836 to 470). We believe this is because the model tends to fit the disordered point clouds into the average vehicle, so the MMD will not have many deviations. In addition, we believe that the MMD is related to the memory characteristics of the model itself, that is, whether it will fit an average car shape for chaotic point clouds rather than diverge. Therefore, different models perform differently for the same batch of input data. In general, however, the MMD experiments verify that we can distinguish the point clouds that do not comply with the completion rules and minimize the errors in the completion process.
   
   \subsection{Time Consumption}
   An essential requirement of our network is to run in real-time. Therefore, we conduct detailed experiments to test the running time of our network.
   
   We test the impact of different input and batch sizes on the network running time. In the experiment, we choose batch sizes 1, 4, 8, 16, and 32 and input sizes 128, 256, 512, 1024, and 2048. We conduct all our experiments on the same GTX 1080 Ti GPU. From the evaluation results in Table II\ref{time}, it is not difficult to see that our network can meet the real-time requirement, whether we evaluate objects one by one or set a large batch for batch evaluation. 
   
   \section{Downstream Application}
   We also test whether our quality evaluation model can help on actual tasks. Following the pipeline of \cite{Najibi2020DOPSLT}, we design a complete process: We use the SOTA detection method to perform detection on the real datasets, obtain the preliminary ROI results, insert the completion model after the ROI for completed point clouds, and add the point clouds to refine the detection result. We insert our evaluation before the completion module to see how much our method can help filter out bad point clouds. All models are pre-trained on separate datasets \textbf{without} any joint optimization. 
   	
	\begin{figure}[ht]
		\centering
		\framebox{\parbox{3.3in}{
				\includegraphics[scale=0.5]{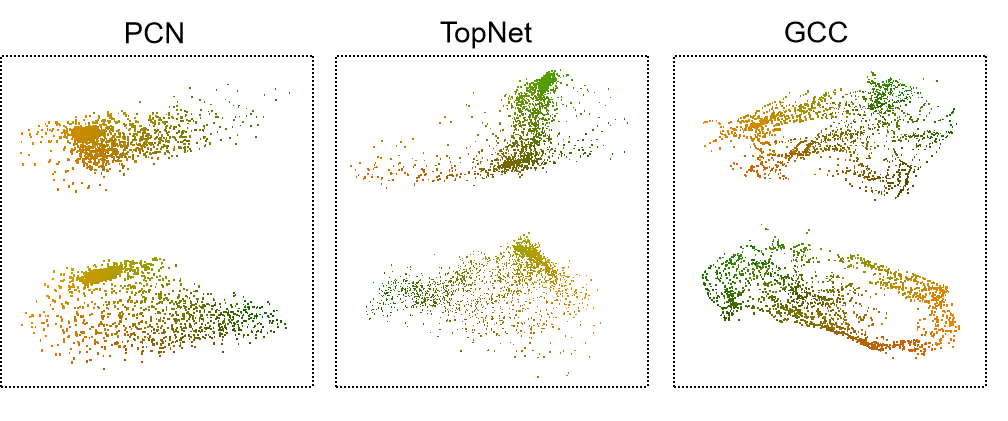}
		}}
		\caption{Cases of completion failure. All input point clouds are point clouds of Cars in KITTI. The wrong completion can not help subsequent tasks and will even disrupt the original pattern of point clouds and introduce a large amount of noise.}
		\label{fail}
      \vspace{-1em}
	\end{figure}
   
    We select Kitti Object as our test set. Since our training data are created based on the PV-RCNN model trained on KITTI, we hope to avoid over-fitting due to the consistency of the model. Therefore, in addition to PV-RCNN, we also conduct comparative experiments using the open-sourced Point-RCNN model on the KITTI dataset. We first use the detection model to obtain the point clouds of vehicles and normalize and rotate them to the forward coordinate system.
   
     We first go through the whole KITTI Object dataset and get an average car number of around 4 per frame. Considering that over-completion may bring about unexpected noise, such as the replicated completion for two close ROIs that contain the same car, we choose to use the 10 ROIs with the highest confidence as the candidate. We then conduct three groups of experiments according to ROI selection strategies. The first is to complete only ROIs that pass the evaluation network. We feed the 10 candidates to the evaluation network, and then get an average of 2 objects passing the evaluation per frame. The second method is to select boxes purely according to the ROI confidence score. Considering the average car number and the average completed car number of the first experiment, we select the top 5 and top 2 boxes for completion. The third ROI selection methods is to select the ROIs that contain the most points. We also conduct the top-5 and top-2 experiments separately. Moreover, we find that that dense point clouds of the completed car may lead to some problems of the network's sampling strategy. We therefore sample 256 points for each object using the furthest point sampling strategy and concatenate them together with the original input point cloud.
   
   From Table \ref{detection_kitti}, we can find that the test results improve over direct completion. This shows that our quality evaluation network can select point clouds suitable for processing and illustrates the importance of the evaluation module.
	
\section{Conclusion}
In this paper, we propose a quality evaluation network that can effectively evaluate whether the real point clouds in actual tasks are suitable for completion. In addition, we design a dataset construction and quality scoring method to minimize the need for manual point cloud evaluation and complete weakly supervised training of the network. After training with synthetic datasets, our network can fully meet the requirements of actual tasks and assist refinement or improvement of SLAM, detection, tracking, and other downstream tasks. Through experiments on synthetic and real datasets, our network shows its rapid, simple, and robust characteristics and practical value.
\balance
\newpage
\bibliographystyle{IEEEtran}
\bibliography{root}
\end{document}